\documentclass[10pt,twocolumn,letterpaper]{article}

\usepackage{cvpr}
\usepackage{times}
\usepackage{epsfig}
\usepackage{graphicx}
\usepackage{amsmath}
\usepackage{amssymb}

\usepackage{dsfont}
\usepackage{booktabs}
\usepackage{pifont}
\usepackage{arydshln}
\usepackage{multirow}
\usepackage[ruled]{algorithm2e}
\SetKwComment{Comment}{$\triangleright$\ }{}

\usepackage{xcolor}

\usepackage{soul}

\DeclareMathOperator*{\argmax}{arg\,max}

\usepackage[pagebackref=true,breaklinks=true,letterpaper=true,colorlinks,bookmarks=false]{hyperref}

\cvprfinalcopy 

\def\cvprPaperID{10219} 

\ifcvprfinal\fi
\begin{document}

\title{Understanding Adversarial Examples from the Mutual Influence of Images and Perturbations}

\author{Chaoning Zhang$^{*}$\\
{\tt\small chaoningzhang1990@gmail.com}
\and
Philipp Benz$^{*}$\\
{\tt\small pbenz@kaist.ac.kr}
\and Tooba Imtiaz\\
{\tt\small timtiaz@kaist.ac.kr}
\and In-So Kweon\\
{\tt\small iskweon@kaist.ac.kr}\\
{\small $^*$ indicates equal contribution }\\
Robotics and Computer Vision (RCV) Laboratory\\
Korea Advanced Institute of Science and Technology (KAIST)\\
291 Daehak-ro, Yuseong-gu, Daejeon 34141, Korea\\
}

\maketitle

\begin{abstract}
A wide variety of works have explored the reason for the existence of adversarial examples, but there is no consensus on the explanation. We propose to treat the DNN logits as a vector for feature representation, and exploit them to analyze the mutual influence of two independent inputs based on the Pearson correlation coefficient (PCC). We utilize this vector representation to understand adversarial examples by disentangling the clean images and adversarial perturbations, and analyze their influence on each other. Our results suggest a new perspective towards the relationship between images and universal perturbations: Universal perturbations contain dominant features, and images behave like noise to them. This feature perspective leads to a new method for generating targeted universal adversarial perturbations using random source images. We are the first to achieve the challenging task of a targeted universal attack without utilizing original training data. Our approach using a proxy dataset achieves comparable performance to the state-of-the-art baselines which utilize the original training dataset. 
\end{abstract}

\section{Introduction}

\begin{figure}[t]
    \centering
    \includegraphics[width=\linewidth]{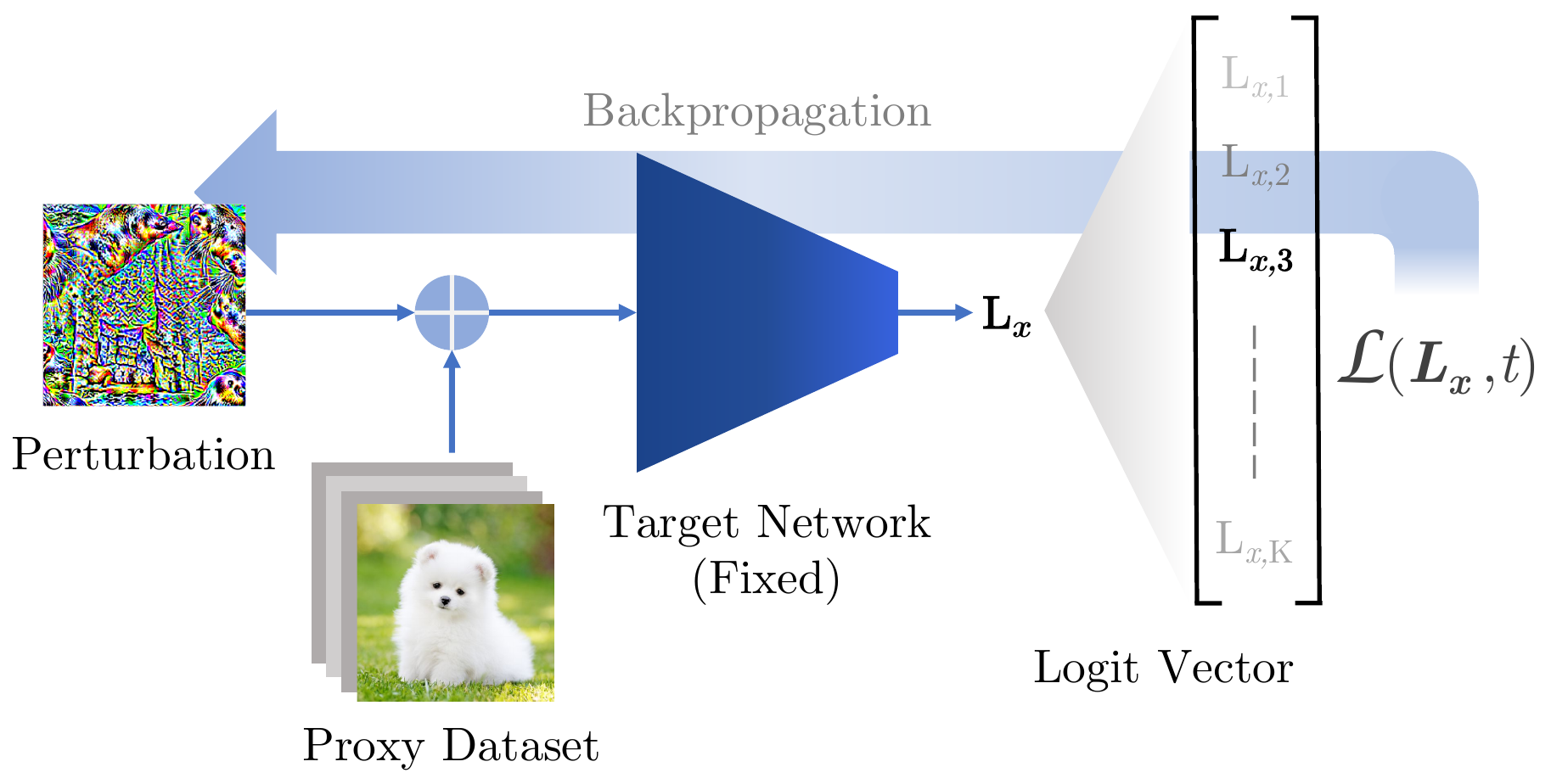}
    \caption{Based on our observation that \emph{adversarial perturbations contain dominant features and images behave like noise to them}, we design a new method of generating targeted universal adversarial perturbations without data, by using a proxy dataset.}
\label{fig:teaser}
\end{figure}

Deep neural networks (DNNs) have shown impressive performance in numerous applications, ranging from image classification~\cite{he2016identity,zhang2019revisiting} to motion regression~\cite{dosovitskiy2015flownet,zhang2020deepptz}. However, DNNs are also known to be vulnerable to adversarial attacks~\cite{szegedy2013intriguing,ranjan2019attacking}. A wide variety of previous works~\cite{goodfellow2014explaining,tabacof2016exploring,tanay2016boundary,koh2017understanding,nakkiran2019a,athalye2018obfuscated} explore the reason for the existence of adversarial examples, but there is a lack of consensus on the explanation~\cite{akhtar2018threat}.
While the working mechanism of DNNs is not fully understood, one widely accepted interpretation considers DNNs as feature extractors~\cite{he2016identity}, which inspires the recent work~\cite{ilyas2019adversarial} to link the existence of adversarial examples to non-robust features in the training dataset. 

Contrary to previous works analyzing adversarial examples as a whole (summation of image and perturbation), we instead propose to analyze adversarial examples by disentangling image and perturbations and studying their mutual influence. Specifically, we analyze the influence of two independent inputs on each other in terms of contributing to the obtained feature representation when the inputs are combined. We treat the network logit outputs as a means of feature representation. Traditionally, only the most important logit values, such as the highest logit value for classification tasks, are considered while other values are disregarded. We propose that all logit values contribute to the feature representation and therefore treat them as a logit vector.
We utilize the Pearson correlation coefficient (PCC)~\cite{anderson2003introduction} to analyze the extent of linear correlation between logit vectors. 
The PCC values computed between the logit vectors of each independent input and the input combination gives insight on the contribution of the two independent inputs towards the combined feature representation. Our proposed general analysis framework is shown to be useful for analyzing influence of any two independent inputs, such as images, Gaussian noise, perturbations, etc. In this work, we limit the focus on analyzing the influence of image and perturbation in universal attacks. Our findings show that for a universal attack, the adversarial examples (AEs) are strongly correlated to the UAP, while a low correlation is observed between AEs and input images (see \figurename~\ref{fig:img_targeted_uap}). This suggests that for a DNN, UAPs dominate over the clean images in AEs, even though the images are visually more dominant. Treating the DNN as feature extractor, we naturally conclude that the UAP has features that are more dominant compared to the features of the images to attack. Consequently we claim that ``UAPs are features while images behave like noise to them". This is contrary to the general perception that treats the perturbation as noise to images in adversarial examples. Our interpretation thus provides a simple yet intuitive insight on the working of UAPs.

The observation, that images behave like noise to UAPs motivates the use of proxy images to generate targeted UAPs without original training data, as shown in \figurename~\ref{fig:teaser}. Our proposed approach is more practical because the training data is generally inaccessible to the attacker~\cite{Mopuri2017datafree}.
Our contributions can be summarized as follows:
\begin{itemize}
    \item We propose to treat the DNN logits as a vector for feature representation. These logit vectors can be used to analyze the contribution of features of two independent inputs when summed towards the output. In particular, our analysis results regarding universal attacks reveal that in an AE, the UAP has dominant features, while the image behaves like noise to them.
    \item We leverage this insight to derive a method using random source images as proxy dataset to generate targeted UAPs without original training data. To our best knowledge, we are the first to fulfill this challenging task while achieving comparable performance to the state-of-the-art baselines utilizing the original training dataset.
\end{itemize}

\section{Related Work}
\label{sec:related_work}
We summarize previous works with two focuses: (1) explanations of adversarial vulnerability and (2) existing adversarial attack methods.

\textbf{Explanation of adversarial vulnerability.}
Goodfellow~\etal attribute the reason of adversarial examples to the local linearity of DNNs, and support their claim by their proposed simple yet effective FGSM~\cite{goodfellow2014explaining}. However, this linearity hypothesis is not fully compatible with the existence of adversarial examples which violate local linearity~\cite{madry2017towards}. Moreover, it can not fully explain the phenomenon that greater robustness is not observed in less linear classifiers~\cite{athalye2018obfuscated,tabacof2016exploring,tanay2016boundary}. Another body of works attributes the reason for low adversarial robustness to high-dimensional input properties~\cite{shafahi2018adversarial,fawzi2018adversarial,mahloujifar2019curse,gilmer2018adversarial}. However, reasonably robust DNNs of high-dimensional inputs can be trained in practice~\cite{madry2017towards,raghunathan2018certified}. One recent work~\cite{ilyas2019adversarial} attributes the reason for the existence of adversarial examples to non-robust features in the dataset. Some previous explanations, ranging from limited training data induced over-fitting~\cite{schmidt2018adversarially,tanay2016boundary} to robustness under noise~\cite{fawzi2016robustness,ford2019adversarial,cohen2019certified}, are well aligned with their framework~\cite{ilyas2019adversarial}. The concept of non-robust features is also implicitly explored in other works~\cite{bubeck2018adversarial,nakkiran2019a}. On the other hand, possible reasons for vulnerability against universal adversarial perturbations have been explored in~\cite{moosavi2017universal, moosavi2017analysis, jetley2018friends, moosavi2018robustness}. Their analysis is mainly based on the network decision boundaries, in particular, the existence of universal perturbations is linked to the large curvature of decision boundary. 
Our work mainly focuses on the explanation of universal adversarial vulnerability. One core aspect that differentiates our analysis framework from previous works is that we explore the influence of images and perturbations on each other, while previous works mainly analyze adversarial example as a whole~\cite{moosavi2017universal, moosavi2017analysis, jetley2018friends}. 
We explicitly analyze how the image and perturbations influence each other. Our analysis framework is mainly based on the proposed logit vector interpretation of how DNNs respond to the features in the input, without relying on the curvature property of decision boundaries~\cite{moosavi2017universal,moosavi2017analysis,jetley2018friends}. 

\textbf{Existing adversarial attack methods.}
The existing attacks are commonly categorized under image-dependent attacks~\cite{szegedy2013intriguing, goodfellow2014explaining, kurakin2016adversarial, moosavi2016deepfool, carlini2017towards} and universal (\ie image-agnostic) attacks~\cite{moosavi2017universal, khrulkov2018art, Mopuri2017datafree, metzen2017universal, poursaeed2018generative, zhang2019cd-uap, naseer2019cross} which devise one single perturbation to attack most images. Image-dependent attack techniques have been explored in a variety of works ranging from optimization based techniques~\cite{szegedy2013intriguing, carlini2017towards} to FGSM related techniques~\cite{goodfellow2014explaining, kurakin2016adversarial, dong2018boosting, wu2018understanding}. Universal adversarial perturbations (UAPs) were first proposed by~\cite{moosavi2017universal}, and deploy the DeepFool attack~\cite{moosavi2016deepfool} iteratively on single data samples. Due to the nature of being image-agnostic, universal attacks constitute a more challenging task than image-dependent ones. 

Another way to categorize attacks is non-targeted vs. targeted attacks. Generative targeted universal perturbations have been explored by~\cite{poursaeed2018generative}. Targeted attacks can be seen as a special, but more challenging case of non-targeted attacks. Class discriminative (CD) UAPs were proposed in~\cite{zhang2019cd-uap}, aiming to fool only a subset of classes. 
The above mentioned universal attacks require utilization of the original training data. However, in practice the attacker often has no access to the training data~\cite{Mopuri2017datafree}. To overcome this limitation, Mopuri~\etal propose to generate universal perturbation without training data~\cite{Mopuri2017datafree}. However, their approach is specifically designed for non-targeted attacks by maximizing the activation scores in every layer, and their performance is inferior to approaches with access to original training data. Another attempt for data-free non-targeted universal attack by training a network to generate proxy images is explored in~\cite{reddy2018ask} . 
No prior work is found to have achieved targeted universal attack without access to the original training data, and our work is the first attempt in this direction.

\section{Analysis Framework}
\subsection{Logit Vector}\label{sec:logit_vector}
Following the common consensus that DNNs are feature extractors, we intend to analyze adversarial examples from the feature perspective.
The logit values are often used as an indicator of feature presence in an image. 
Previous works~\cite{jetley2018friends, ilyas2019adversarial}, however, mainly focus only on the DNN highest logit output indicating the predicted class, while all other logits are usually neglected. ``Logits" refer to the DNN output before the final softmax layer. 
In this work, we assume that all DNN output logit values represent the network response to features in the input. One concern about this vector interpretation is that only the logits of the ground-truth classes or other semantically similar classes are meaningful, while the other logits might be just random (small) values and thus do not carry important information. We address this concern after introducing the terms and notation used throughout this work.

A deep classifier $\hat{C}$ maps an input image $x \in \mathds{R}^d$ with a pixel range of $[0,1]$ to an output logit vector $L_{x}=\hat{C}(x)$. The vector $L_{x}$ has $K$ entries corresponding to the total number of classes. The predicted class $y_x$ of an input $x$ can then be calculated from the logit vector as $y_x=\argmax(L_{x})$. We adopt the logit vector to facilitate the analysis of the mutual influence of two independent inputs in terms of their contribution to the combined feature representation. We mainly consider two independent inputs $a \in \mathds{R}^d$ and $b \in \mathds{R}^d$, which can be images, Gaussian noise, perturbations, etc., whose corresponding logit vectors are denoted as $L_{a}$ and $L_{b}$, respectively. The summation of these two inputs $c = a + b$, when fed to a DNN, leads to the feature representation $L_c$. Both inputs $a$ and $b$ contribute partially to $L_c$. Moreover, it is reasonable to expect that the contribution of each input will be influenced by the other one. Specifically, the extent of influence will be reflected in the linear correlation between the individual logit vector $L_{a}$ (or $L_{b}$) and $L_c$.

\subsection{Pearson Correlation Coefficient}
In statistics, the Pearson correlation coefficient (PCC)~\cite{anderson2003introduction} is a widely adopted metric to measure the linear correlation between two variables. In general, this coefficient is defined as
\begin{equation}
    \text{PCC}_{X,Y} = \frac{cov(X, Y)}{\sigma_{X} \sigma_{Y}},
\end{equation}
where $cov$ indicates the covariance and $\sigma_{X}$ and $\sigma_{Y}$ are the standard deviation of vector $X$ and $Y$, respectively, and the PCC values range from $-1$ to $1$. The absolute value indicates the extent to which the two variables are linearly correlated, with $1$ indicating perfect linear correlation, $0$ indicating zero linear correlation, and the sign indicates whether they are positively or negatively correlated. Treating the logit vector as a variable, the PCC between different logit vectors can be calculated. We are mainly concerned about $\text{PCC}_{L_a, L_c}$ and $\text{PCC}_{L_b, L_c}$, since $\text{PCC}_{L_a, L_b}$ is always close to zero due to independence. Comparing $\text{PCC}_{L_a, L_c}$ and $\text{PCC}_{L_b, L_c}$ can provide insight about the contribution of the two inputs to $L_c$, with a higher PCC value indicating the more significant contributor. For example, if $\text{PCC}_{{L_a, L_c}}$ is larger than $\text{PCC}_{L_b, L_c}$, input $a$'s share can be seen as more dominant than input $b$ towards the final feature response. The relationship of two logit vectors, $L_{a}$ and $L_{c}$ for instance, can be visualized by plotting each logit pair. The extent of their correlation can be observed and quantified by the PCC. 
\begin{figure}[t]
    \centering
    \includegraphics[width=\linewidth]{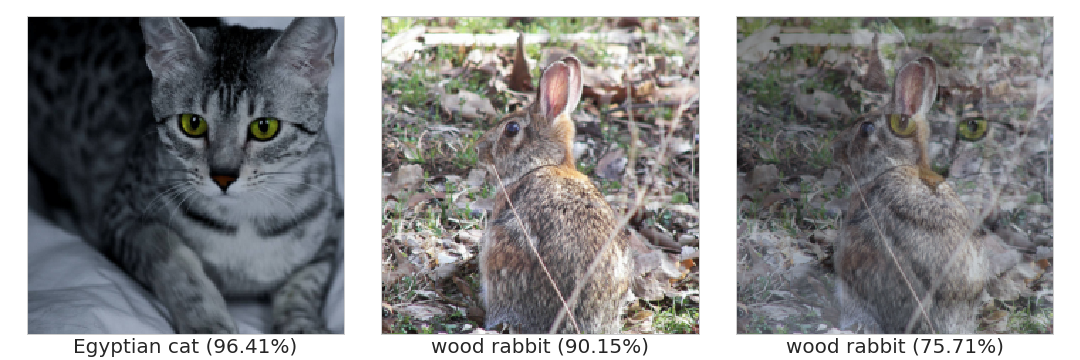}
    \includegraphics[width=\linewidth]{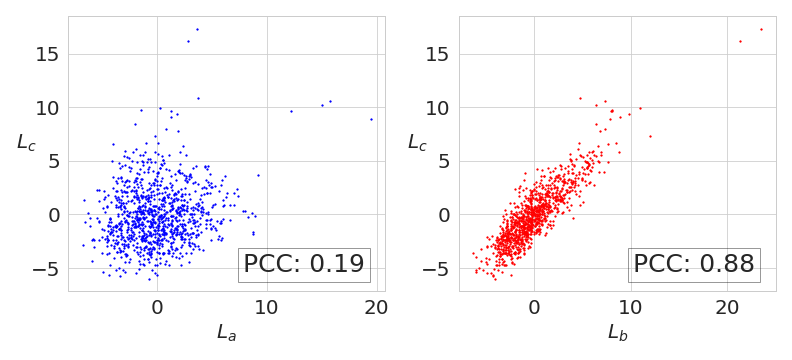}
    \caption{Images and their logit vector analysis. The first row shows the sample images $a$ and $b$ and the resulting image $c$. The second row shows the plots of logit vector $L_c$ over $L_a$ (left) and $L_b$ (right), with their respective PCC values.}
    \label{fig:sample_analysis}
\end{figure}

As a basic example, we show the logit vector analysis of two randomly sampled images from ImageNet~\cite{krizhevsky2012imagenet} in \figurename~\ref{fig:sample_analysis}. 
The plot shows a strong linear correlation between $L_b$ and $L_c$ ($\text{PCC}_{L_b, L_c}=0.88$), while $L_a$ and $L_c$ are practically uncorrelated ($\text{PCC}_{L_a, L_c}=0.19$). 
These observations suggest a dominant contribution of input $b$ towards logit vector $L_c$. As a result, the same label ``Wood rabbit" is predicted for $c$ and $b$. Such combination of images has also been explored in Mixup~\cite{zhang2018mixup} for training classifiers.

\begin{table}[h]
\centering
\small
\setlength\tabcolsep{1.5pt}
    \caption{PCC analysis for VGG19 using $1000$ image pairs randomly sampled from the ImageNet test set. Here, for each image pair, the mean and standard deviations of higher and lower PCC values are reported under $\text{PCC}_h$ and $\text{PCC}_l$, respectively.}
    \begin{tabular}{ccccccc}
    \toprule
           & $|S|$  & $\text{PCC}_h$         & $\text{PCC}_l$         & $\text{PCC}_h - \text{PCC}_l$ & $\mathcal{P}_{\text{PCC}}$ \\
    \midrule
    $S_m$ & $445$ & $0.74 \pm 0.10$ & $0.27 \pm 0.23$ & $0.47 \pm 0.27$ & $96\%$ \\
    $S_n$ & $555$ & $0.63 \pm 0.13$ & $0.33 \pm 0.20$ & $0.30 \pm 0.22$ & -      \\
    \bottomrule
    \end{tabular}
\label{tab:PCC_metric_analysis}
\end{table}
To establish the reliability of the PCC value as a metric, we repeat the above experiment with $1000$ image pairs and report results on the effectiveness of PCC to predict label $c$ in Table~\ref{tab:PCC_metric_analysis}.
We divide the image pairs into two groups: $S_m$ and $S_n$. $S_m$ comprises of image pairs having the same predicted class $y_c$ as the prediction $y_a$ or $y_b$. For $S_n$, the predicted class $y_c$ is different from both $y_a$ and $y_b$. Moreover, we use the parameter $\mathcal{P}_{\text{PCC}}$ to show the proportion of predictions correctly inferred from the PCC values relative to the network predictions for $c$. For the image pairs from set $S_m$, the $\mathcal{P}_{\text{PCC}}$ is $96\%$, confirming the reliability of the PCC as our metric.
The high gap between $\text{PCC}_h$ and $\text{PCC}_l$ further provides evidence for the high $\mathcal{P}_{\text{PCC}}$.
For the image pairs from $S_n$, $\text{PCC}_h - \text{PCC}_l$ is smaller, implying that neither of the inputs is significantly dominant.

Recall that there is a concern that most logit values might be just random values, which is partially addressed by observing the correlation between PCC and $y_c$ as shown in \figurename~\ref{fig:sample_analysis}. If the concern were valid, such that only a few logits are meaningful (\ie only the highest logits or the logits for semantically similar classes), a high divergence should be observed for the less significant logits. However, this assumption does not align well with the results in \figurename~\ref{fig:sample_analysis}, thus confirming the importance of all logit values. A higher PCC value for the dominant input further rules out the concern that the lower logit values are random. 

\section{Influence of Images and Perturbations on Each Other}
In this section, we analyze the interaction of clean images with Gaussian noise perturbation, universal perturbations and image-dependent perturbations. In doing so, input $a$ is the image and input $b$ the perturbation. The analysis is performed on VGG19 pretrained on ImageNet. For consistency, a randomly chosen $a$ (shown in \figurename~\ref{fig:sample_analysis}, top left) is used for all experiments. Along the same lines, for targeted perturbations we randomly set `sea lion' as the target class $t$. For more results with different images and target classes on different networks, please refer to the supplementary material.

\subsection{Analysis of Gaussian Noise}
\begin{figure}[t]
    \centering
    \includegraphics[width=\linewidth]{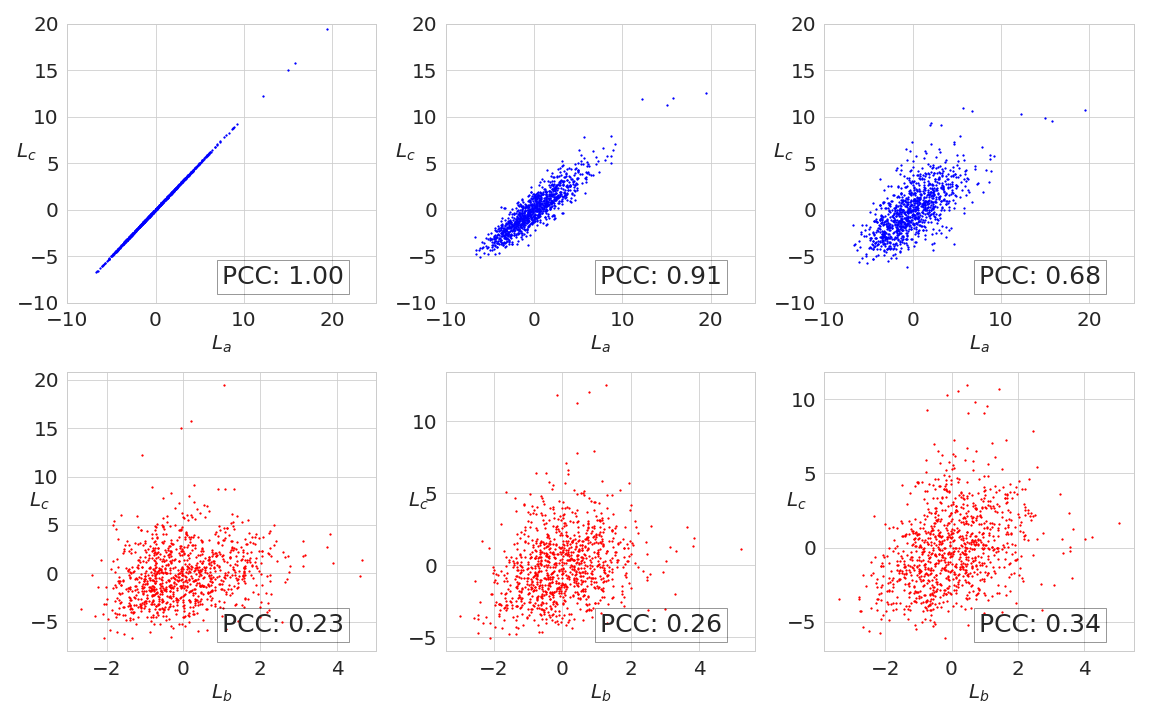}
    \caption{Logit vector analysis for an input image and Gaussian noise $\mathcal{N(\mu, \sigma)}$. The analysis is shown for $\mu=0$ and $\sigma=0$ (left), $\sigma=0.1$ (middle) and $\sigma=0.2$ (right))}
    \label{fig:analysis_various_noise}
\end{figure}
To facilitate the interpretation of our main experiment of performing analysis for perturbations, we first show the influence of noise (Gaussian noise) on images. The Gaussian noise is sampled from $\mathcal{N(\mu, \sigma)}$ with $\mu=0$ and different standard deviations. The relationship between $L_{a}$, $L_{c}$ is visualized in \figurename~\ref{fig:analysis_various_noise}. As expected, by adding zero magnitude Gaussian noise (\ie no Gaussian noise) to the image, $L_{a}$ and $L_{c}$ are perfectly linearly correlated ($\text{PCC}_{L_a, L_c}=1$). If the Gaussian noise magnitude is increased ($\sigma=0.1$ for instance), $L_{a}$ and $L_{c}$ still show a high linear correlation ($\text{PCC}_{L_a, L_c}=0.91$).
Investigating the relationship between $L_b$ and $L_c$, a low correlation can be observed for all noise inputs $b$ indicating a low contribution to the final prediction. 

\subsection{Analysis of Universal Perturbations}
\begin{figure}[t]
    \centering
    \includegraphics[width=\linewidth]{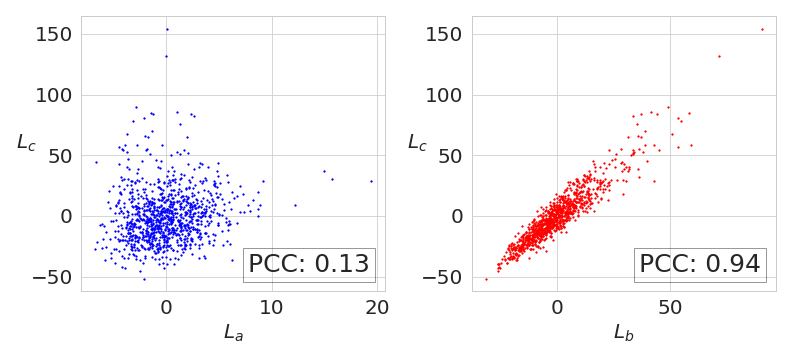}
    \caption{Logit vector analysis for input image ($a$) and targeted UAP ($b$). The targeted UAP was trained for target class `sea lion' and loss function $\mathcal{L}^{t}_{CL2}$}
    \label{fig:img_targeted_uap}
\end{figure}
\begin{figure}[t]
    \centering
    \includegraphics[width=\linewidth]{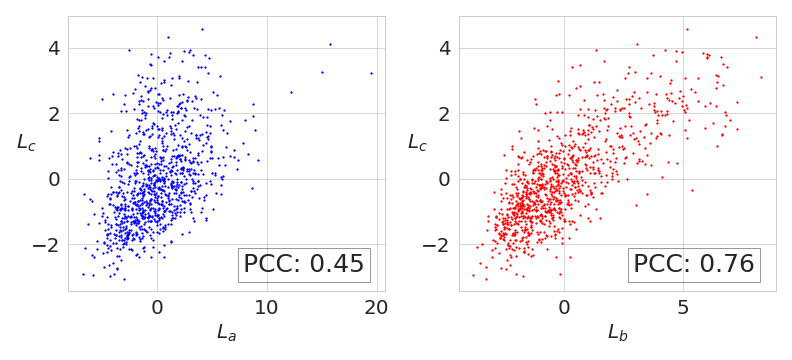}
    \caption{Logit vector analysis for input image ($a$) and non-targeted UAP ($b$). The UAP was trained with loss function Equation~\ref{eq:non_targeted_loss}}
    \label{fig:img_non_targeted_uap}
\end{figure}

Universal perturbations come in two flavors: targeted and non-targeted. We use Algorithm~\ref{alg:uap} with loss function $\mathcal{L}^{t}_{CL2}$ to generate targeted universal perturbations, and generate non-targeted universal perturbations using Equation~\ref{eq:non_targeted_loss} as the loss function. The results of this analysis are shown for a targeted and non-targeted UAP in \figurename~\ref{fig:img_targeted_uap} and \figurename~\ref{fig:img_non_targeted_uap}, respectively. For the targeted scenario, two major observations can be made: First, $\text{PCC}_{L_{a},L_{c}}$ is smaller than $\text{PCC}_{L_{b},L_{c}}$, indicating a higher linear correlation between $L_{c}$ and $L_{b}$ than $L_{c}$ and $L_{a}$. In other words, the features of the perturbation are more dominant than that of the clean image. Second, $\text{PCC}_{L_{a},L_{c}}$ is close to $0$, indicating that the influence of the perturbation on the image is so significant that the clean image features are seemingly unrecognizable to the DNN. 
In fact, comparing the logit analysis of $L_a$ and $L_c$ in \figurename~\ref{fig:img_targeted_uap} with that of Gaussian noise and image in \figurename~\ref{fig:analysis_various_noise} (bottom), a striking similarity is observed. 
This offers a novel interpretation of targeted universal perturbations:
\textbf{Targeted universal perturbations themselves (independent of the images to attack) are features, while images behave like noise to them.} 
We further explore the non-targeted perturbations, and report the results in \figurename~\ref{fig:img_non_targeted_uap}. Similar to targeted universal perturbations, the $\text{PCC}_{L_{a},L_{c}}$ is smaller than $\text{PCC}_{L_{b},L_{c}}$ for the non-targeted perturbation. However the dominance of the non-targeted perturbation is not as significant as that of the targeted perturbation. 

\subsection{Analysis of Image-Dependent Perturbations}
\begin{figure}[t]
    \centering
    \includegraphics[width=\linewidth]{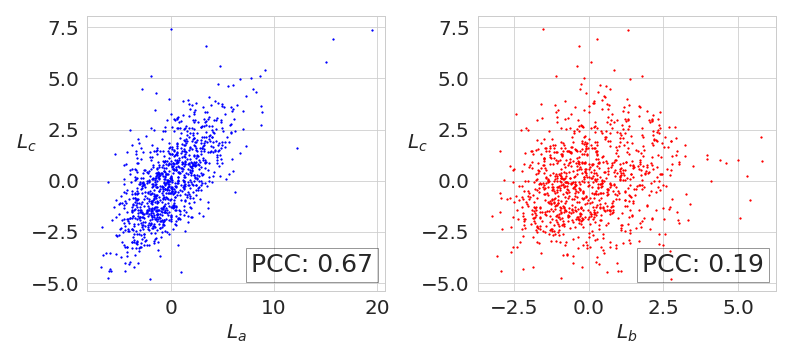}
    \caption{Logit vector analysis for input image ($a$) and targeted image-dependent perturbation ($b$). The perturbation was crafted with PGD~\cite{madry2017towards}, with target class `sea lion'}
    \label{fig:targeted_img_dependent}
\end{figure}
\begin{figure}[t]
    \centering
    \includegraphics[width=\linewidth]{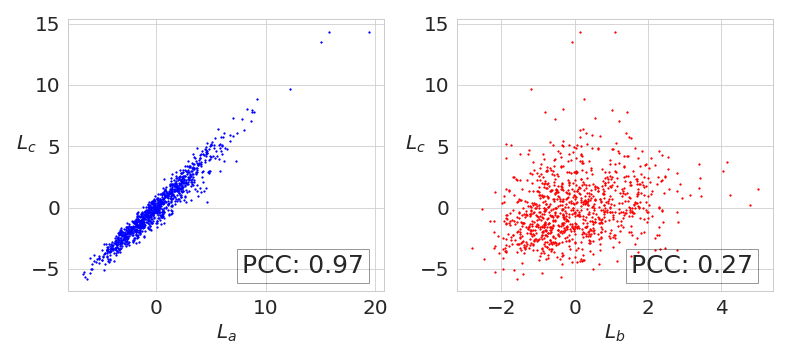}
    \caption{Logit vector analysis for input image ($a$) and non-targeted image-dependent perturbation ($b$). The perturbation was crafted with PGD~\cite{madry2017towards}}
    \label{fig:non_targeted_img_dependent}
\end{figure}
The logit vector analysis results for targeted and non-targeted image-dependent perturbations are reported in \figurename~\ref{fig:targeted_img_dependent} and \figurename~\ref{fig:non_targeted_img_dependent}, respectively. Contrary to the universal perturbations, the image-dependent perturbations are weakly correlated to $c$, and have a noise-like behaviour (\figurename~\ref{fig:analysis_various_noise}).
However, the image gets misclassified even though the image features appear to be more dominant than the perturbation. This is because the image features are more strongly corrupted through the image-dependent perturbation than Gaussian noise. This special behavior appears due to the fact that the image-dependent perturbations are crafted to form concrete features only in combination with the image. Such image-dependent behavior violates our assumption of independent inputs. However, we include these results since they offer additional insight into adversarial examples. 

\subsection{Why Do Adversarial Perturbations Exist?} \label{sec:why_do}
A wide variety of works have explored the existence of adversarial examples as discussed in section~\ref{sec:related_work}. 
Based on our previous analyses, we arrive at the following explanation for the existence of UAPs:

Universal adversarial perturbations contain features independent of the images to attack. The image features are corrupted to an extent of being unrecognizable to a DNN, and thus the input images behave like noise to the perturbation features. 

The finding in~\cite{jetley2018friends} that universal perturbations behave like features of a certain class aligns well with our statement. Jetley~\etal argue that universal perturbations exploit the high-curvature image-space directions to behave like features, while our finding suggests that universal perturbations themselves contain features independent of the images to attack.
Utilizing the perspective of positive curvatures of decision boundaries, Jetley~\etal adopt the decision boundary-based attack DeepFool~\cite{moosavi2016deepfool}. 
However, our explanation does not explicitly rely on the decision boundary properties, but focuses on the occurrences of strong features, robust to the influence of images. We can therefore deploy the PGD algorithm to generate perturbations consisting of target class features similar to~\cite{ilyas2019adversarial}. 

If universal perturbations themselves contain features independent of the images to attack, do image-dependent perturbations behave in a similar way?
As previously discussed, the analysis results in \figurename~\ref{fig:targeted_img_dependent} reveal that the behavior of image-dependent perturbations is not like features, but noise. On the other hand, the original image features are retained to a high extent. 
Ilyas~\etal~\cite{ilyas2019adversarial} revealed that image-dependent adversarial examples include the features of the target class. However, as seen from the analysis in subsection~\ref{sec:why_do}, the isolated perturbation seems not to retain independent features due its low PCC value, but rather interacts with the image to form the adversarial features.  

\section{Targeted UAP with Proxy Data}
\label{sec:targeted_uap_with_proxy_data}
Our above analysis demonstrates that images behave like noise to the universal perturbation features. Since the images are treated like noise, we can exploit proxy images as background noise to generate targeted UAPs without the original training data. The proxy images do not need to have any class object belonging to the original training class and their main role is to make the targeted UAP have strong background-robust target class features. 
\subsection{Problem Definition}
Formally, given a data distribution $\mathcal{X} \in \mathds{R}^d$ of images, we compute a single perturbation vector $v$ that satisfies
\begin{align}
    \begin{split}
        &\hat{C}(x+v) = t \quad \text{for \textit{most} } x \sim \mathcal{X} \\
        &||v||_p \leq \epsilon.
    \end{split}
    \label{eq:objective}
\end{align}
The magnitude of $v$ is constrained by $\epsilon$ to be imperceptible to humans. $||\cdot||_p $ refers to the $l_p$-norm and in this work, we set $p=\infty$ and $\epsilon = 10$ for images in range $[0, 255]$\footnote{For images in the range $[0, 1]$, $\epsilon=\frac{10}{255}$} as in~\cite{moosavi2017universal}. Specifically, we assume having no access to original training data. Thus, the training data $\mathcal{X}_{v}$ for $v$ generation can be different from the original dataset $\mathcal{X}$. We denote the proxy dataset as $\mathcal{X}_{v}$.

To evaluate targeted UAPs, we use the targeted fooling ratio metric~\cite{poursaeed2018generative}, \ie the ratio of samples fooled into the target class to the number of all data samples. We also use the non-targeted fooling ratio~\cite{poursaeed2018generative, moosavi2017universal}, calculating the ratio of misclassified samples to the total number of samples, for evaluation. 

\subsection{Loss Function and Algorithm}
\begin{algorithm}[t]
    \SetAlgoLined
    \DontPrintSemicolon
    \SetKwInput{KwInput}{Input}
    \SetKwInput{KwOutput}{Output}
    \SetKwFunction{FOptim}{Optim}
    \KwInput{Proxy data $\mathcal{X}_v$, Classifier $\hat{C}$, Loss function $\mathcal{L}$, mini-batch size $m$, Number of iterations $I$, perturbation magnitude $\epsilon$}
    \KwOutput{Perturbation vector $v$}
    $v \leftarrow 0$ \Comment*[r]{Initialize}
    \For {iteration $=1, \dots, I$}{
        $B \sim \mathcal{X}_v$: $|B| =  m$ \Comment*[r]{Randomly sample}
        $g_v \leftarrow \underset{x \sim B}{\mathds{E}} [\nabla_{v} \mathcal{L}$] \Comment*[r]{Calculate gradient} 
        $v \leftarrow$ \FOptim{$g_v$} \Comment*[r]{Update} 
        $v \leftarrow \epsilon \frac{v}{||v||_p}$ \Comment*[r]{Norm projection}
        }
\caption{UAP algorithm}
\label{alg:uap}
\end{algorithm}

To achieve the desired objective Eq.~\ref{eq:objective} most naively, the commonly used cross-entropy loss function $\mathcal{L}_{\text{CE}}$ can be utilized. Since cross-entropy loss holistically incorporates logits of all classes, this loss function leads to overall lower fooling ratios. This behavior can be resolved by using a loss function $\mathcal{L}_{\text{L}}$ that only aims to increase the logit of the target class. 

Since we consider universal perturbations, to balance the above objective between different samples in training, we extend $\mathcal{L}_{\text{L}}$ by clamping the logit values as follows:
\begin{equation}
\label{eq:targeted_loss}
    \mathcal{L}^{t}_{CL1} = \max(\underset{i \neq t}\max \hat{C}_i(x_v + v) - \hat{C}_{t}(x_v + v) , -\kappa)
\end{equation}
where $\kappa$ indicates the confidence value, $x_v$ are samples from the proxy data $\mathcal{X}_v$ and $\hat{C}_i$ indicates the $i$-th entry of the logit vector. In this case, the proxy data can be either a random source dataset or the original training data, depending on data availability. Note that similar techniques of clamping the logits have also been used in~\cite{carlini2017towards}, however, their motivation is to obtain minimum-magnitude (image-dependent) perturbations. While the target logit in loss function $\mathcal{L}^{t}_{CL1}$ is increased, the logit values of $\max \hat{C}_i(x_v + v)$ are decreased simultaneously during the training process. This effect is undesirable for generating a UAP with strong target class features, since other classes except the target classes will be included in the optimization, which might have negative effects on the gradient update. 
To prevent manipulation of logits other than the target class, we exclude the non-targeted class logit values in the optimization step, such that these values are only used as a reference value for clamping the target class logit.
We indicate this loss function as $\mathcal{L}^{t}_{CL2}$. We report an ablation study of the different loss function performances in Table~\ref{tab:loss_ablation}. The results suggest that $\mathcal{L}^{t}_{CL2}$, in general, outperforms all other discussed loss functions.
We further provide a loss function resembling $\mathcal{L}^{t}_{CL2}$ for the generation of non-targeted UAPs.
\begin{equation}
\label{eq:non_targeted_loss}
    \mathcal{L}^{nt} = \max(\hat{C}_{gt}(x_v + v) - \underset{i \neq gt}\max \hat{C}_i(x_v + v), -\kappa)
\end{equation}
In the special case of crafting non-targeted UAPs, the proxy dataset has to be the original training dataset.

\begin{table*}[t]
\centering
\caption{Ablation study on the performance of different loss functions, for the proposed targeted UAP. The values in each column represent mean and standard deviation of the non-targeted fooling ratio ($\%$) and targeted fooling ratio ($\%$) obtained for $5$ runs and target class `sea lion'.}
\label{tab:loss_ablation}
    \small
    \setlength\tabcolsep{3.pt}
    \scalebox{0.95}{
    \begin{tabular}{c|cc|cc|cc|cc|cc}
        \toprule
        Loss & \multicolumn{2}{c}{AlexNet}  & \multicolumn{2}{c}{GoogleNet} & \multicolumn{2}{c}{VGG16} & \multicolumn{2}{c}{VGG19} & \multicolumn{2}{c}{ResNet152} \\
        \midrule
        $\mathcal{L}_{\text{CE}}$   & $\mathbf{90.5 \pm 0.6}$ & $55.4 \pm 1.0$  & $70.8 \pm 1.5$ & $55.2 \pm 2.2$  & $89.1 \pm 0.3$ & $75.9 \pm 0.9$  & $87.9 \pm 0.5$ & $70.8 \pm 1.1$  & $78.2 \pm 0.9$ & $66.5 \pm 1.3$ \\
        $\mathcal{L}_{\text{L}}$   & $89.2 \pm 0.4$ & $47.1 \pm 1.1$  & $71.6 \pm 0.8$ & $56.9 \pm 1.1$  & $91.0 \pm 0.3$ & $79.0 \pm 0.6$  & $90.8 \pm 0.2$ & $73.1 \pm 0.8$  & $80.1 \pm 0.8$ & $69.1 \pm 0.4$ \\
        $\mathcal{L}^{t}_{CL1}$    & $90.2 \pm 0.3$ & $\mathbf{57.6 \pm 1.4}$ & $71.7 \pm 1.4$ & $57.9 \pm 2.3$  & $90.1 \pm 0.4$ & $80.3 \pm 0.5$  & $88.2 \pm 0.3$ & $\mathbf{75.5 \pm 0.6}$  & $80.2 \pm 0.3$ & $\mathbf{71.4 \pm 0.5}$ \\
        $\mathcal{L}^{t}_{CL2}$ & $\mathbf{90.5 \pm 0.3}$ & $49.4 \pm 1.2$  & $\mathbf{73.0 \pm 1.5}$ & $\mathbf{58.4 \pm 2.2}$  & $\mathbf{93.5 \pm 0.3}$ & $\mathbf{82.8 \pm 0.7}$  & $\mathbf{92.7 \pm 0.1}$ & $72.3 \pm 2.5$  & $\mathbf{81.3 \pm 1.1}$ & $70.6 \pm 2.1$ \\
        \bottomrule
    \end{tabular}
    }
\end{table*}

We provide a simple, yet effective algorithm in Algorithm~\ref{alg:uap}. Our gradient based method adopts the ADAM~\cite{kingma2014adam} optimizer and mini-batch training, which have also been adopted in the context of data-free universal adversarial perturbations~\cite{reddy2018ask}. Mopuri~\etal train a generator network for crafting UAPs with this configurations, which can be considered more complex. 

\subsection{Main Results}

\begin{table*}[t]
\centering
\caption{Results for targeted UAPs trained on four different datasets. The values in each column represent mean and standard deviation of the non-targeted fooling ratio ($\%$) and targeted fooling ratio ($\%$) obtained for $8$ different target classes.}
\label{tab:targeted_uap_four_datasets}
    \small
    \setlength\tabcolsep{3.pt}
    \scalebox{0.95}{
    \begin{tabular}{c|cc|cc|cc|cc|cc}
        \toprule
        Proxy Data & \multicolumn{2}{c}{AlexNet}  & \multicolumn{2}{c}{GoogleNet} & \multicolumn{2}{c}{VGG16} & \multicolumn{2}{c}{VGG19} & \multicolumn{2}{c}{ResNet152} \\
        \midrule
        ImageNet~\cite{krizhevsky2012imagenet}  
        & $89.9 \pm 2.2$ & $48.6 \pm 13.3$ & $77.7 \pm 3.2$ & $59.9 \pm 6.6$ & $92.5 \pm 1.3$ & $75.0 \pm 7.8$  & $91.6 \pm 1.3$ & $71.6 \pm 6.9$  & $80.8 \pm 2.6$ & $66.3 \pm 7.0$ \\
        COCO~\cite{lin2014microsoft}
        & $89.9 \pm 2.6$ & $47.2 \pm 13.1$ & $76.8 \pm 3.7$ & $59.8 \pm 7.5$ & $92.2 \pm 1.7$ & $75.1 \pm 12.3$ & $91.6 \pm 1.5$ & $68.8 \pm 9.4$  & $79.9 \pm 2.9$ & $65.7 \pm 7.8$ \\
        VOC~\cite{Everingham10}
        & $88.9 \pm 2.6$ & $46.9 \pm 12.7$ & $76.7 \pm 3.2$ & $58.9 \pm 6.0$ & $92.2 \pm 1.6$ & $74.7 \pm 7.9$  & $90.5 \pm 2.3$ & $68.8 \pm 8.2$  & $79.1 \pm 3.3$ & $65.2 \pm 7.1$ \\
        Places365~\cite{zhou2017places} 
        & $90.0 \pm 2.1$ & $42.6 \pm 16.4$ & $76.4 \pm 3.7$ & $60.0 \pm 5.4$ & $92.1 \pm 1.5$ & $73.4 \pm 9.6$  & $91.5 \pm 1.6$ & $64.5 \pm 17.0$ & $78.0 \pm 3.2$ & $62.5 \pm 9.9$ \\
        \bottomrule
    \end{tabular}
    }
    \label{alg:targeted_uaps_various_datasets}
\end{table*}

\begin{table}[t]
\centering
\caption{Comparison of the proposed method to other methods. The results are divided in universal attacks with access to the original ImageNet training data (upper) and data-free methods (lower). The metric is reported in the non-targeted fooling ratio ($\%$))}
\label{tab:uap}
    \small
    \setlength\tabcolsep{1.2pt}
    \scalebox{0.9}{
    \begin{tabular}{cccccc}
        \toprule
        Method  & AlexNet & GoogleNet  & VGG16  & VGG19  & ResNet152 \\
        \midrule
        UAP~\cite{moosavi2017universal}       & $93.3$           & $78.9$           & $78.3$           & $77.8$           & $84.0$           \\
        GAP~\cite{poursaeed2018generative}    &  -               &  $82.7$          & $83.7$           & $80.1$           &     -       \\
        Ours(ImageNet)                        & $\mathbf{96.17}$ & $\mathbf{88.94}$ & $\mathbf{94.30}$ & $\mathbf{94.98}$ & $\mathbf{90.08}$ \\
        \midrule
        FFF~\cite{Mopuri2017datafree}         & $80.92$          & $56.44$          & $47.10$          & $43.62$          & -                \\
        AAA~\cite{reddy2018ask}               & $89.04$          & $75.28$          & $71.59$          & $72.84$          & $60.72$          \\
        GD-UAP~\cite{mopuri2018generalizable} & $87.02$          & $71.44$          & $63.08$          & $64.67$          & $37.3$           \\
        Ours (COCO)                           & $\mathbf{89.9}$ & $\mathbf{76.8}$ & $\mathbf{92.2}$ & $\mathbf{91.6}$ & $\mathbf{79.9}$ \\
        \bottomrule
    \end{tabular}
    }
    \label{tab:performance_comparison}
\end{table}

\begin{table}[t]
\centering
\caption{Transferability results for the proposed targeted universal adversarial attack. The attack was performed for target class `sea lion' and proxy dataset MS-COCO. The rows indicate the source model and the columns indicates the target model. The values in each column are reported in the non-targeted fooling ratio ($\%$) and targeted fooling ratio ($\%$)}
\label{tab:transfer_ratio}
    \small
    \setlength\tabcolsep{1.0pt}
    \scalebox{0.77}{
    \begin{tabular}{c cc cc cc cc cc}
        \toprule
                    & \multicolumn{2}{c}{AlexNet} & \multicolumn{2}{c}{GoogleNet}  & \multicolumn{2}{c}{VGG-16}  &  \multicolumn{2}{c}{VGG19}  &  \multicolumn{2}{c}{ResNet152} \\
        \midrule
        AlexNet     & $\mathbf{90.45}$ & $\mathbf{49.61}$ & $54.77$ & $0.01$  & $60.43$ & $0.13$  & $58.66$ & $0.09$  & $47.02$ & $0.02$  \\
        GoogleNet   & $53.25$ & $0.02$  & $\mathbf{75.47}$ & $\mathbf{62.06}$ & $50.51$ & $0.17$  & $48.79$ & $0.14$  & $34.94$ & $0.34$  \\
        VGG16       & $53.71$ & $0.03$  & $41.26$ & $0.02$  & $\mathbf{93.62}$ & $\mathbf{82.90}$ & $82.99$ & $13.69$ & $36.73$ & $0.01$  \\
        VGG19       & $53.67$ & $0.02$  & $39.78$ & $0.02$  & $83.40$ & $44.53$ & $\mathbf{92.53}$ & $\mathbf{75.61}$ & $35.36$ & $0.01$  \\
        ResNet152   & $54.46$ & $0.03$  & $42.43$ & $0.07$  & $55.05$ & $1.63$  & $55.12$ & $1.05$  & $\mathbf{80.47}$ & $\mathbf{70.20}$ \\
        \bottomrule
    \end{tabular}
    }
    \label{tab:transferability}
\end{table}

\begin{table}[t]
\centering
\caption{Results for Transferability measured with PCC values. Generated with COCO as background, for target class sea lion. The rows indicate the source model and the columns indicates the target model.}
\label{tab:transfer_pcc}
    \small
    \setlength\tabcolsep{1.2pt}
    \scalebox{0.9}{
    \begin{tabular}{c cc cc cc cc cc}
        \toprule
                    & AlexNet & GoogleNet  & VGG-16  &  VGG19  & ResNet152 \\
        \midrule 
        AlexNet     & $\mathbf{1.00}$ & $0.09$ & $0.24$ & $0.14$ & $-0.05$  \\
        GoogleNet   & $0.24$ & $\mathbf{1.00}$ & $0.24$ & $0.14$ & $0.00$   \\
        VGG16       & $0.36$ & $0.09$ & $\mathbf{1.00}$ & $0.48$ & $-0.11$ \\
        VGG19       & $0.19$ & $0.07$ & $0.55$ & $\mathbf{1.00}$ & $-0.09$  \\
        ResNet152   & $0.28$ & $0.11$ & $0.36$ & $0.30$ & $\mathbf{1.00}$  \\
        \bottomrule
    \end{tabular}
    }
\end{table}

We generate the targeted UAPs for four different datasets, the ImageNet training set as well as three proxy datasets. In Algorithm~\ref{alg:uap}, we set the number of iterations to $1000$, use loss function $\mathcal{L}^{t}_{CL2}$ and a learning rate of $0.005$ with batch-size $32$. As the proxy datasets, we use images from MS-COCO~\cite{lin2014microsoft} and Pascal VOC~\cite{Everingham10}, two widely used object detection datasets, and Places365~\cite{zhou2017places}, a large-scale scene recognition dataset. 
We generated targeted UAPs with the $4$ datasets for $8$ different target classes and evaluate them on the ImageNet test dataset. The average over the $8$ target scenarios are reported in Table~\ref{alg:targeted_uaps_various_datasets}. Two major observations can be made: First, a significant difference can not be observed for the three different proxy datasets. Moreover, there is only a marginal performance gap between training with the proxy datasets and training with the original ImageNet training data. The results support our assumption that the influence of the input images on targeted UAPs is like noise. 

We also explored generating targeted UAPs with white images and Gaussian noise as the proxy dataset. In both scenarios, inferior performance was observed. We refer the reader to the supplementary material for a discussion about possible reasons and further results. 

\begin{figure}[t]
    \centering
    \includegraphics[width=\linewidth]{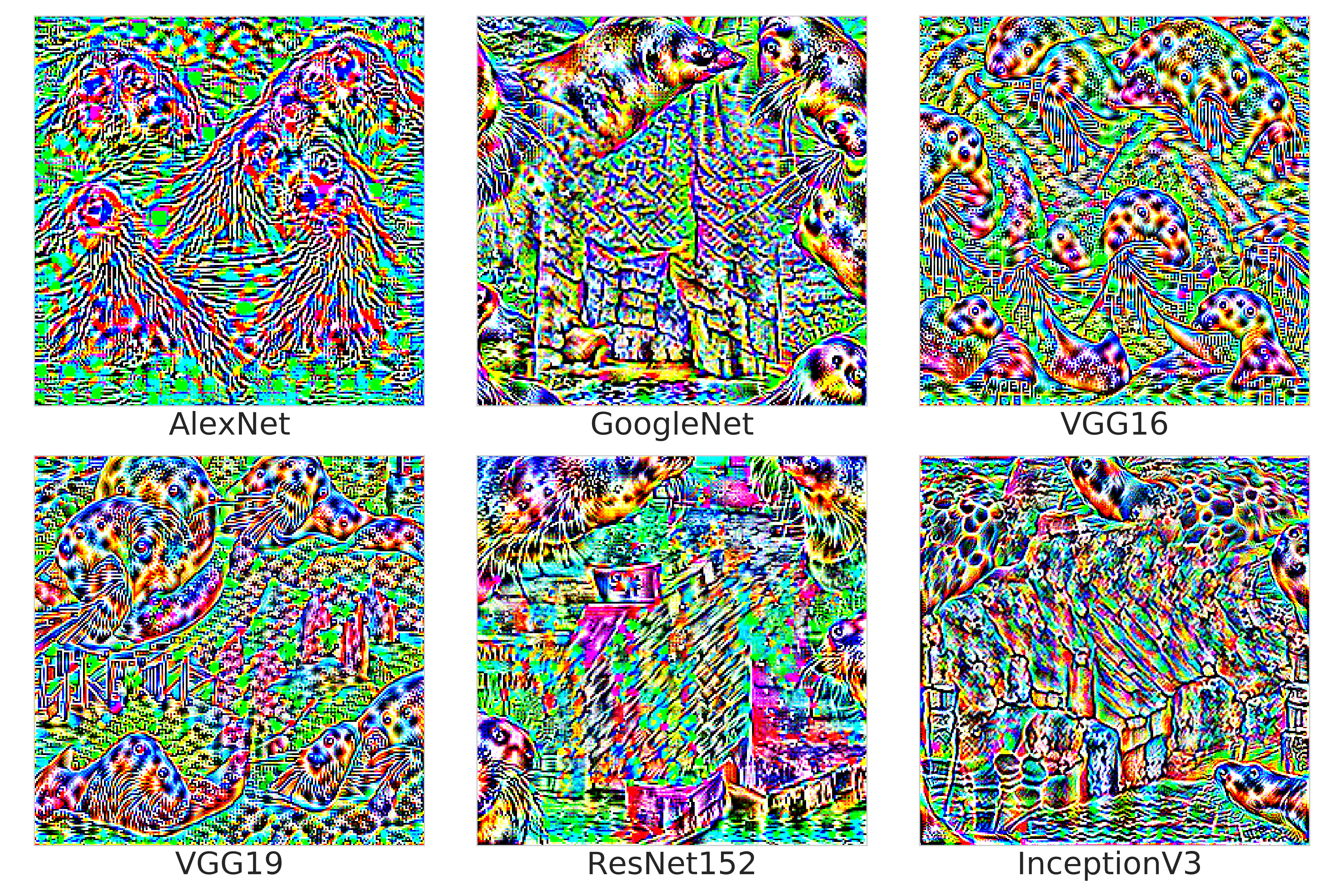}
    \caption{Targeted universal perturbations (target class `sea lion') for different network architectures.}
    \label{fig:perturbations_qual}
\end{figure}

\begin{figure}[t]
    \centering
    \includegraphics[width=\linewidth]{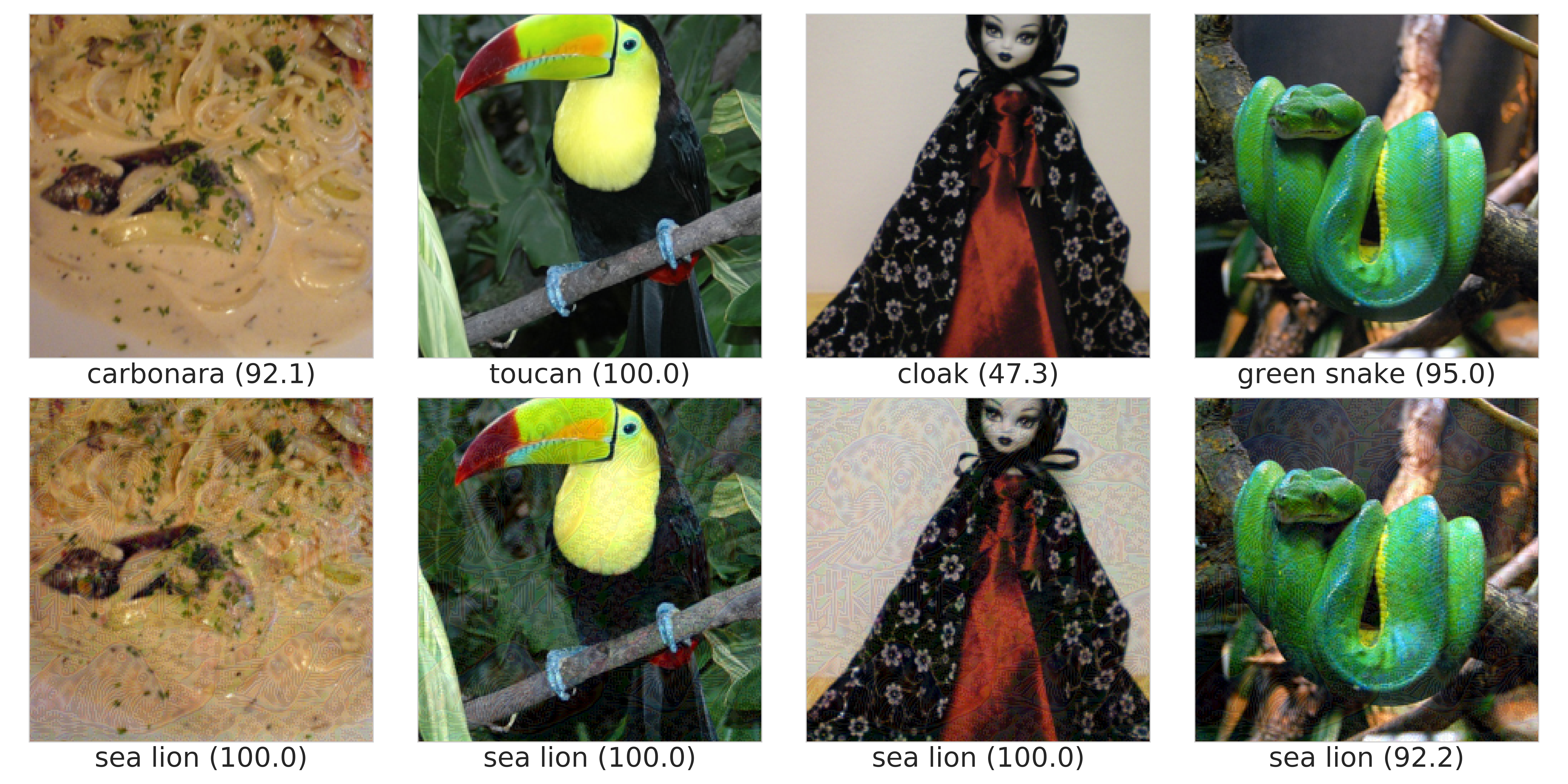}
    \caption{Qualitative Results. Clean images (top) and perturbed images (bottom) for VGG19}
    \label{fig:images_qual}
\end{figure}

Targeted perturbations for different networks are shown in \figurename~\ref{fig:perturbations_qual}. Since the target class is sea lion, we can notice the existence of sea lion-like patterns by taking a closer look. 
Samples of clean images and perturbed images misclassified as sea lion are shown in \figurename~\ref{fig:images_qual}.  

\subsection{Comparison with Previous Methods}
To the best of our knowledge, this is the first work to achieve targeted UAP without original training data, thus we can only compare our performance with previous works on related tasks. The authors of~\cite{poursaeed2018generative} report a targeted fooling ratio of $52\%$ for Inception-V3 with access to the ImageNet training dataset. We use COCO as the proxy dataset and achieve a superior performance of $53.4\%$.
We can not find any other targeted UAP method available in the literature but other previous works report the (non-targeted) fooling ratio and we compare our performance with them and the results are available in Table~\ref{tab:performance_comparison}. We distinguish between methods with and without data availability. To compare with the methods with data-availability we trained a non-targeted UAP on ImageNet utilizing our introduced non-targeted loss function from Equation~\ref{eq:non_targeted_loss}. Note that we do not block the gradient for $\underset{i \neq gt}\max \hat{C}_i(x_v + v)$ to let the algorithm automatically search a dominant class for an effective attack. We observe that our approach achieves superior performance than both UAP~\cite{moosavi2017universal} and GAP~\cite{poursaeed2018generative}. For the case without access to the original training dataset, we use the COCO dataset to generate the UAP, and report the averages of performance on $8$ target classes. Note that our method still generates a targeted UAP, but we use the non-targeted metric for performance evaluation. This setting is in favor of other methods, since ideally, we could report the best performance of a certain target class. Without bells and whistles, our method achieves comparable performance to the state-of-the-art data-free methods, constituting evidence that our simple approach is efficient. 



\subsection{Transferability}
The transferability results are available in Table~\ref{tab:transfer_ratio}. We observe that the non-targeted transferability performs reasonably well, while targeted transferability does not. We find no previous work reporting the targeted transferability for universal perturbations. For image-dependent perturbations, the targeted transferability has been explored in~\cite{han2019once}, which reveals that the targeted transferability is unsatisfactory when source network and target network belong to different network families. When the networks belong to the same network family, relatively higher transferability can be observed~\cite{han2019once}. This aligns well with our finding that VGG16 and VGG19 transfer reasonably well between each other as presented in Table~\ref{tab:transferability}. We further report the PCC of the two network UAPs in Table~\ref{tab:transfer_pcc}. We observe that the PCC values are relatively higher between VGG16 and VGG19 than other networks, indicating an additional benefit of PCC to provide insight to network transferability. 



\section{Conclusion}
In this work, we treat the DNN logit output as a vector to analyze the influence of two independent inputs in terms of contributing to the combined feature representation.
Specifically, we demonstrate that the Pearson correlation coefficient (PCC) can be used to analyze relative contribution and dominance of each input.
Under the proposed analysis framework, we analyze adversarial examples by disentangling images and perturbations to explore their mutual influence. Our analysis results reveal that universal perturbations have dominant features and the images to attack behave like noise them. This new insight yields a simple yet effective algorithm, with a carefully designed loss function, to generate targeted UAPs by exploiting a proxy dataset instead of the original training data. We are the first to achieve this challenging task and the performance is comparable to state-of-the-art baselines utilizing the original training dataset. 

\section{Acknowledgement}
We thank Francois Rameau and Dawit Mureja Argaw for their comments and suggestions throughout this project. This work was supported by NAVER LABS and the Institute for Information \& Communications Technology Promotion (2017-0-01772) grant funded by the Korea government.


{\small
\bibliographystyle{ieee_fullname}
\bibliography{egbib}
}
\end{document}